# ARTICLE INFORMATION

**Article title**

BaitBuster-Bangla: A Comprehensive Dataset for Clickbait Detection in Bangla with Multi-Feature and Multi-Modal Analysis

**Authors**

Abdullah Al Imran[a,*], Md Sakib Hossain Shovon[a], M. F. Mridha[a]

**Affiliations**

[a]Advanced Machine Intelligence Research Lab (AMIRL), Dhaka, Bangladesh.

**Corresponding author's email address and Twitter handle**

abdalimran@gmail.com (Abdullah Al Imran)

**Abstract**

This study presents a large multi-modal Bangla YouTube clickbait dataset consisting of 253,070 data points collected through an automated process using the YouTube API and Python web automation frameworks. The dataset contains 18 diverse features categorized into metadata, primary content, engagement statistics, and labels for individual videos from 58 Bangla YouTube channels. A rigorous preprocessing step has been applied to denoise, deduplicate, and remove bias from the features, ensuring unbiased and reliable analysis. As the largest and most robust clickbait corpus in Bangla to date, this dataset provides significant value for natural language processing and data science researchers seeking to advance modeling of clickbait phenomena in low-resource languages. Its multi-modal nature allows for comprehensive analyses of clickbait across content, user interactions, and linguistic dimensions to develop more sophisticated detection methods with cross-linguistic applications.

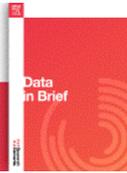

# SPECIFICATIONS TABLE

| | |
|---|---|
| **Subject** | Computer Science, Data Science |
| **Specific subject area** | Applied Machine Learning, Natural Language Processing |
| **Data format** | Processed, Labeled, Analyzed |
| **Type of data** | Text, Table |
| **Data collection** | The dataset was collected using the official YouTube API for data retrieval. The collection process involved making a curated list of 28 Not Clickbait, and 26 Clickbait Bangla YouTube channels, and querying the API with specific search parameters to retrieve a set of videos in the Bengali language. The data collection process was automated using Python web automation libraries. The collected dataset was then labeled and preprocessed to remove bias from the titles and descriptions, ensuring a fair evaluation on downstream tasks. |
| **Data source location** | YouTube (https://www.youtube.com/) |
| **Data accessibility** | Repository name: Mendeley Data<br>Data identification number: 10.17632/3c6ztw5nft.1<br>Direct URL to data:<br>https://data.mendeley.com/datasets/3c6ztw5nft/ |

# VALUE OF THE DATA

This dataset holds significant value for the scientific community and can benefit various stakeholders in the field of computer science and natural language processing who want to research low resource languages like Bangla. Here are several reasons why these data are valuable:

- The dataset is the biggest multi-feature and multi-modal dataset in the Bangla language to date, offering a valuable resource for investigating clickbait in the context of Bangla video content sharing.

- The dataset is unique because it includes a broad set of features, like video metadata, user engagement data, and thumbnail image URLs. This multi-modal data would enable researchers to perform comprehensive analyses and develop more sophisticated clickbait detection algorithms.

- The dataset underwent a rigorous debiasing and noise removal process, enhancing its reliability and usability. It also includes three types of labels, such as auto labels, human labels, and AI model labels, making it versatile for various research methodologies.

- The dataset allows for comparative analysis between different languages or regions, shedding light on similarities, differences, and cultural nuances in clickbait creation and user engagement patterns. This sheds light on universal and language-specific motivations and strategies, furthering overall understanding of the phenomenon.

- The dataset can be leveraged to explore new research directions, such as analyzing the impact of clickbait on user engagement metrics, investigating the effectiveness of countermeasures against clickbait, or studying the evolution of clickbait techniques over time. It also supports socio-cultural analysis of online content dynamics within the Bangla community.

- The dataset's compatibility with similar datasets in other languages allows for the development of multilingual clickbait detection models. Researchers can combine this dataset with others to create models capable of identifying clickbait across different linguistic contexts, contributing to cross-linguistic clickbait research and detection efforts.

# DATA DESCRIPTION

The dataset contains a total of 253,070 records, with 18 features. The features are categorized into four different types: Metadata, Primary Data, Engagement Stats, and Label. Under the Metadata category contains basic information about the channel and video, such as their unique identifiers, date and time of publication, and thumbnail URLs. The Primary Data category contains information about the title and description of the video. The "Processed" columns refer to the cleaned data after denoising, deduplication and debiased for further analysis. The Engagement Stats category contains data on user engagement metrics for each video. The Label category contains predefined auto labels, human annotated labels, and AI generated pseudo labels. Table 1 presents a detailed overview and definitions of the features.

Table 1: Detailed overview and definitions of the features

| Feature Type | Feature Name | Data Type | Definition |
| --- | --- | --- | --- |
| Metadata | channel_id | string | ID of the YouTube channel |
| Metadata | channel_name | string | Name of the YouTube channel |
| Metadata | channel_url | string | URL of the YouTube channel |
| Metadata | video_id | string | ID of the video |
| Metadata | publishedAt | datetime | Date and time when the video was published |
| Primary Data | title | string | Title of the video |
| Primary Data | title_debiased | string | Debiased title of the video |
| Primary Data (Processed) | description | string | Debiased description of the video |
| Primary Data (Processed) | description_debiased | string | Description of the YouTube video without bias |
| Metadata | url | string | URL of the video |
| Engagement Stats | viewCount | int | Number of views the video has received |
| Engagement Stats | commentCount | int | Number of comments on the video |
| Engagement Stats | likeCount | int | Number of likes on the video |
| Engagement Stats | dislikeCount | int | Number of dislikes on the video |
| Metadata | thumbnails | string | URL of the thumbnail for the video |
| Label | auto_labeled | string | Automatically labeled using predefined rule |
| Label (Processed) | human_labeled | string | Labeled by human |
| Label (Processed) | ai_labeled | string | Labeled by an AI model fine-tuned on human labeled data |

# EXPERIMENTAL DESIGN, MATERIALS AND METHODS

Our experimental design, Fig. 1, consists of five stages - Collection, Standardization, Labeling, AI-based Labeling, and preparing the Final Dataset (BaitBuster-Bangla).

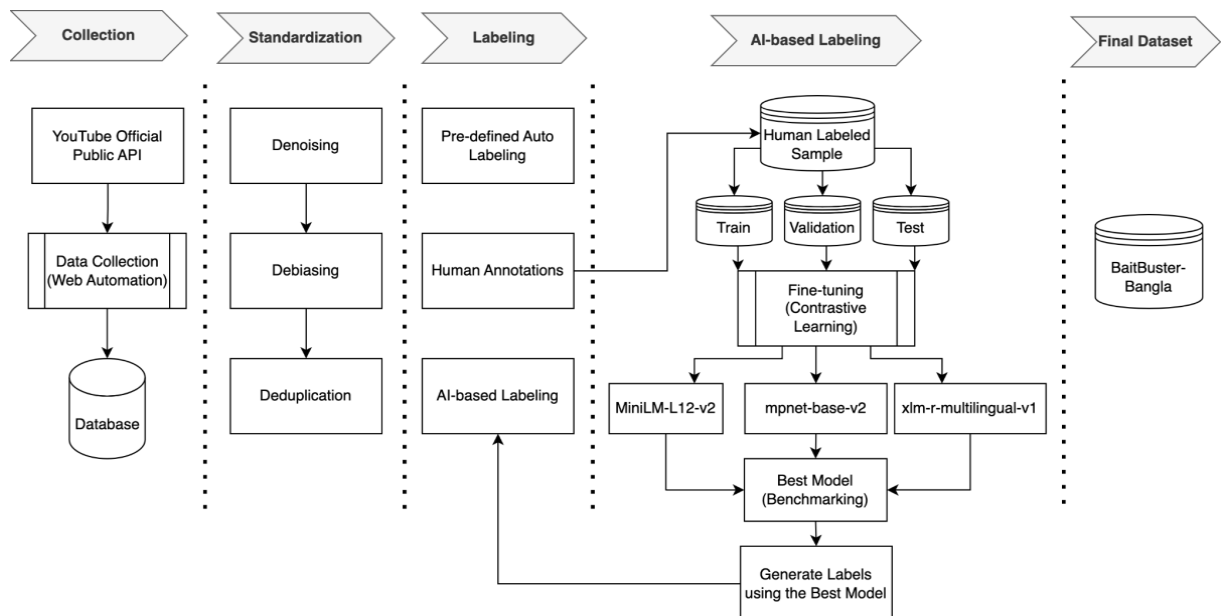

Fig. 1: Experimental design flow

# 1. Collection

For data collection, we utilized the YouTube official public API to access metadata and textual features of videos from 54 popular Bangla video sharing channels. We developed an automatic data collection framework leveraging web automation libraries in Python like Selenium, Requests and BeautifulSoup4. This allowed us to iteratively collect all available data from each channel in a scalable manner. The raw, unprocessed data was stored in partitioned Parquet format for efficient querying and manipulation of the large, multimodal dataset containing video metadata, engagement statistics and primary text features.

# 2. Standardization

To ensure data quality, we implemented several standardization steps. Firstly, all HTML tags and other noises were removed from text fields via denoising. We then dropped duplicate data points having identical video titles and descriptions through deduplication. Finally, we performed debiasing using fuzzy string matching to help match and remove identifiable information such as channel names and descriptors from the title and description features to anonymize data. We generated two new columns named title_debiased, and description_debiased. This reduced biases and anomalies in preparing a clean dataset.

# 3. Labeling

Initial labels were assigned to the entire dataset based on predefined channel classifications - 28 channels were labeled as 'Not Clickbait' and 26 as 'Clickbait' through content analysis. Then, a stratified sample of 10,000 data points were manually annotated by volunteer human evaluators to establish a ground truth dataset.

# 4. AI-based Labeling

This human-labeled subset was divided into train, validation, and test splits in a 60:20:20 ratio. We've used only the "title_debiased" column as a feature. Three pretrained multilingual models [4] - MiniLM-L12-v2, mpnet-base-v2 and xlm-r-multilingual-v1 - were fine-tuned using contrastive learning [3] on this dataset. Their performances were benchmarked on the test split. The following table 2 presents the performances of the models on the validation and test dataset. The metrics include Overall ACC (Accuracy), F1 Macro, F1 Micro, and Kappa scores.

Table 2: Performance benchmark on the validation and test dataset

|  | MiniLM-L12-v2 | | mpnet-base-v2 | | xlm-r-multilingual-v1 | |
| --- | --- | --- | --- | --- | --- | --- |
|  | Validation | Test | Validation | Test | Validation | Test |
| **Overall ACC** | 0.982 | 0.985 | 0.991 | 0.988 | 0.990 | 0.990 |
| **F1 Macro** | 0.982 | 0.984 | 0.990 | 0.988 | 0.989 | 0.989 |
| **F1 Micro** | 0.982 | 0.985 | 0.991 | 0.988 | 0.990 | 0.980 |
| **Kappa** | 0.963 | 0.968 | 0.981 | 0.976 | 0.979 | 0.979 |

From the above table we can see, the mpnet-base-v2 model achieves the highest accuracy on the validation set (0.991), followed closely by xlm-r-multilingual-v1 (0.990). However, on the test set, xlm-r-multilingual-v1 achieves the highest accuracy (0.990), while MiniLM-L12-v2 and mpnet-base-v2 have the accuracy of 0.985, and 0.988. Like accuracy, the mpnet-base-v2 model achieves the highest F1 Macro score on the validation set (0.990), followed by xlm-r-multilingual-v1 (0.989). On the test set, xlm-r-multilingual-v1 has the highest F1 Macro score (0.989), while MiniLM-L12-v2 and mpnet-base-v2 have the score of 0.984, and 0.988. The F1 Micro scores align with the accuracy scores. On both the validation and test sets, mpnet-base-v2 achieves the highest F1 Micro score (0.991 and 0.988, respectively). MiniLM-L12-v2 and xlm-r-multilingual-v1 have the same F1 Micro scores (0.982 and 0.980) on the validation and test sets, respectively. The Kappa scores indicate the agreement between the predicted and actual labels, considering chance agreement. The mpnet-base-v2 model achieves the highest Kappa score on both the validation and test sets (0.981 and 0.976). MiniLM-L12-v2 and xlm-r-multilingual-v1 have similar Kappa scores, with MiniLM-L12-v2 being slightly higher on the test set (0.968 vs. 0.979).

According to the above analysis, mpnet-base-v2 consistently performs well across multiple metrics and datasets, making it a strong contender for the best model.

We have also analyzed the performance of the tuned models against the existing Bangla Clickbait dataset [2]. The dataset has 3004 entries with 1560 labeled as clickbait and 1444 labeled as not clickbait. The test performance on this dataset is shown in the following table 3.

Table 3: Performance on existing dataset [2]

| | **Overall ACC** | **F1 Macro** | **F1 Micro** | **Kappa** |
|---|---|---|---|---|
| **MiniLM-L12-v2** | 0.800 | 0.800 | 0.800 | 0.600 |
| **mpnet-base-v2** | 0.805 | 0.805 | 0.805 | 0.610 |
| **xlm-r-multilingual-v1** | 0.787 | 0.787 | 0.787 | 0.573 |

In terms of accuracy, mpnet-base-v2 achieves the highest score with 0.805, followed closely by MiniLM-L12-v2 with 0.800. xlm-r-multilingual-v1 has the lowest accuracy score of 0.787. The F1 Macro scores are consistent with the accuracy scores. mpnet-base-v2 achieves the highest F1 Macro score of 0.805, followed by MiniLM-L12-v2 with 0.800. xlm-r-multilingual-v1 has the lowest F1 Macro score of 0.787. Like accuracy and F1 Macro, mpnet-base-v2 achieves the highest F1 Micro score of 0.805, followed by MiniLM-L12-v2 with 0.800. xlm-r-multilingual-v1 has the lowest F1 Micro score of 0.787. In terms of the Kappa scores, mpnet-base-v2 achieves the highest Kappa score of 0.610, followed by MiniLM-L12-v2 with 0.600. xlm-r-multilingual-v1 has the lowest Kappa score of 0.573.

Considering the overall analysis of the performance metrics, mpnet-base-v2 consistently outperforms the other models in terms of accuracy, F1 Macro, F1 Micro, and Kappa scores. It achieves the highest scores across all metrics, indicating better overall performance. Therefore, based on these results, we've considered mpnet-base-v2 as the best model among the three and used this model to generate the pseudo label, "ai_labeled".

# 5. Final Dataset (BaitBuster-Bangla)

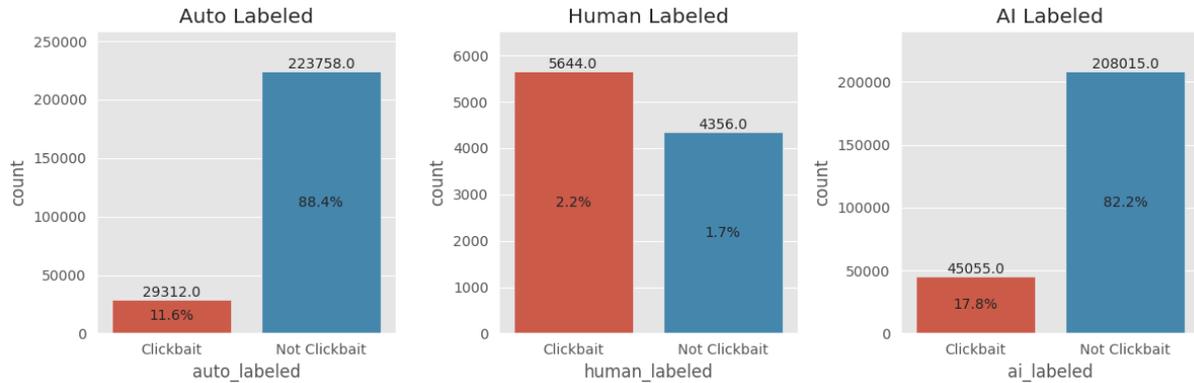

Fig. 2: Distribution of clickbait and non-clickbait entries

The final BaitBuster-Bangla dataset [1] contains 7 metadata features, 2 primary features, 2 processed primary features, 4 user engagement features, and 3 labels including the human annotations and AI generated pseudo labels. Figure 2 presents the distribution of clickbait and non-clickbait entries for all the 3 labels. The final dataset is available in formats such as CSV, parquet and xlsx to facilitate easy analysis and sharing of this resource aimed at tackling clickbait in Bangla online videos.

# ETHICS STATEMENT

The data collected for this study was obtained through an automated process using the YouTube API and Python web automation frameworks. No human subjects were involved in the data collection process, and all data collected is publicly available on the YouTube platform. Therefore, informed consent was not required for this study.

The data does not include any personal information that could be used to identify individuals. All data has been de-identified and anonymized to protect the privacy of users.

The data collection process complied with the terms of service of the YouTube platform. No data was collected that violated the platform's policies.

# CRediT AUTHOR STATEMENT

**Abdullah Al Iman**: Conceptualization, Methodology, Coding, Data curation, Writing, Original draft preparation. **Md Sakib Hossain Shovon**: Writing-Reviewing and Editing. **M. F. Mridha**: Supervision.

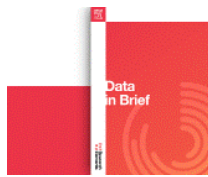

## ACKNOWLEDGEMENTS
This research did not receive any specific grant from funding agencies in the public, commercial, or not-for-profit sectors. Special thanks to Fatema Tuj Johora for volunteering as a human annotator in this project.

## DECLARATION OF COMPETING INTERESTS
The authors declare that they have no known competing financial interests or personal relationships that could have appeared to influence the work reported in this paper.